\documentclass{article}

\usepackage[preprint]{neurips_2026}

\usepackage[utf8]{inputenc}
\usepackage[T1]{fontenc}
\usepackage{hyperref}
\usepackage{url}
\usepackage{booktabs}
\usepackage{amsfonts}
\usepackage{amsmath,amssymb}
\usepackage{nicefrac}
\usepackage{microtype}
\usepackage{xcolor}
\usepackage{graphicx}
\usepackage{multirow}
\usepackage{subcaption}
\usepackage{enumitem}
\usepackage{pifont}
\usepackage{colortbl}

\let\cite\citep

\renewcommand{\acksection}{\section*{Acknowledgments}}

\definecolor{linkblue}{RGB}{76,110,155}
\hypersetup{
    colorlinks=true,
    linkcolor=linkblue,
    citecolor=linkblue,
    urlcolor=linkblue
}
\definecolor{lightblue}{RGB}{235,243,255}
\definecolor{lightgray}{RGB}{245,245,245}
\newcommand{\dui}{\ding{51}}
\newcommand{\budui}{\ding{55}}

\graphicspath{{./figures/}{../figures/}}

\title{Beyond a Single Frame: Multi-Frame Spatially Grounded Reasoning Across Volumetric MRI}

\author{%
  Lama Moukheiber$^{1}$ \quad
  Caleb M.~Yeung$^{2,3}$ \quad
  Haotian Xue$^{1}$ \quad
  Alec Helbling$^{1}$ \quad
  Zelin Zhao$^{1}$ \quad
  Yongxin Chen$^{1}$ \\[0.8em]
  $^{1}$\raisebox{-0.3ex}{\includegraphics[height=2.5ex]{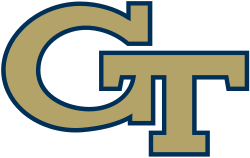}}\,Georgia Institute of Technology \quad
  $^{2}$\raisebox{-0.3ex}{\includegraphics[height=2.5ex]{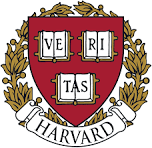}}\,Harvard University \quad
  $^{3}$\raisebox{-0.3ex}{\includegraphics[height=2.5ex]{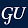}}\,Georgetown University
}

\begin{document}

\maketitle

\begin{center}
    {
        \renewcommand{\arraystretch}{1.2}
        \begin{tabular}{@{} c @{\hskip 24pt} c @{\hskip 24pt} c @{}}
            \href{https://lamawmouk.github.io/SGMRI-VQA}{%
                \raisebox{-3pt}{\includegraphics[height=16pt]{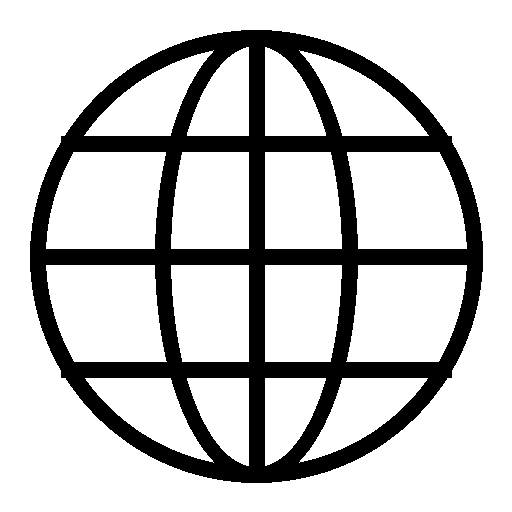}}%
                \hspace{4pt}\textsf{Project Page}%
            } &
            \href{https://huggingface.co/lamamkh/Qwen3-VL-8B-SGMRIVQA}{%
                \raisebox{-3pt}{\includegraphics[height=14pt]{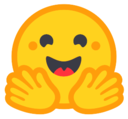}}%
                \hspace{4pt}\textsf{Hugging Face}%
            } &
            \href{https://github.com/lamawmouk/SGMRI-VQA}{%
                \raisebox{-3pt}{\includegraphics[height=14pt]{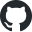}}%
                \hspace{4pt}\textsf{GitHub}%
            }
        \end{tabular}
    }
    \vspace{2mm}
\end{center}

\begin{abstract}
Spatial reasoning and visual grounding are core capabilities for vision-language models (VLMs), yet most medical VLMs produce predictions without transparent reasoning or spatial evidence. Existing benchmarks further evaluate VLMs on isolated 2D images, overlooking the volumetric nature of clinical imaging where findings can span multiple frames or appear on only a few slices. We introduce \textbf{Spatially Grounded MRI Visual Question Answering (SGMRI-VQA)}, a 41,307-pair benchmark for multi-frame spatially grounded reasoning on volumetric MRI. Built from expert radiologist annotations in the fastMRI+ dataset across brain and knee studies, each QA pair pairs a clinician-aligned chain-of-thought trace with frame-indexed bounding-box coordinates, and tasks are organized hierarchically across detection, localization, counting/classification, and captioning---requiring models to jointly reason about what is present, where it is, and across which frames it extends. We benchmark 10 VLMs and show that supervised fine-tuning of Qwen3-VL-8B with bounding-box supervision consistently improves grounding performance over strong zero-shot baselines, indicating that targeted spatial supervision is an effective path toward grounded clinical reasoning. Our code, benchmark, and model outputs are available at \url{https://lamawmouk.github.io/SGMRI-VQA}.
\end{abstract}

\section{Introduction}

Spatial reasoning and visual grounding are core capabilities for vision-language models (VLMs), requiring not only recognition of visual content but also precise localization of findings within complex scenes. Recent VLMs have demonstrated progress on these tasks for natural images---identifying objects, describing their spatial relationships, and producing bounding box coordinates~\cite{bai2025qwen3,chen2024internvl,team2023gemini,achiam2023gpt,munasinghe2025videoglamm}. In the medical domain, VLMs have achieved notable performance on visual question answering benchmarks, combining visual perception with clinical language understanding to identify abnormalities, classify findings, and generate diagnostic descriptions~\cite{zhang2023pmc,hu2024omnimedvqa,ye2024gmai,sun2024pathmmu,bansal2024medmax}. Yet most VLMs remain opaque, offering predictions without the transparent, step-by-step reasoning clinicians rely on---and critically, without spatially grounding their findings. This raises the question: can VLMs \textit{spatially ground} their medical knowledge with chain-of-thought reasoning, localizing findings with precision across complex volumetric scenes?

Currently, existing benchmarks evaluate VLMs on only single 2D images~\cite{liu2025gemex,bercea2025nova,nguyen2025vindr}. However, real-world medical imaging---particularly MRI---is inherently volumetric: a single scan produces dozens to hundreds of sequential slices that together form a continuous 3D representation. Consequently, findings may span multiple frames or appear only on specific slices. MedFrameQA~\cite{yu2025medframeqa} introduced multi-image medical VQA in YouTube videos on modalities including MRI, X-Ray, CT, and radiograph, but with only 2 to 5 temporally related video frames rather than spatially contiguous volumetric slices commonly found in medical scans such as knee and brain MRI. Therefore, limited benchmarks have addressed cross-slice spatial grounding---localizing \textit{what} is present, \textit{where} it is located, and \textit{across which frames} it extends.

To comprehensively evaluate how well VLMs handle multi-frame spatial reasoning across volumetric MRI slices, we introduce Spatially Grounded MRI Visual Question Answering (\textbf{SGMRI-VQA}), a large-scale VQA benchmark for MRI that combines slice-level and volume-level evaluation with volumetric visual grounding and chain-of-thought reasoning across two anatomical domains (brain and knee MRI). Built on expert radiologist annotations from the fastMRI+ dataset~\cite{zhao2022fastmri+} with QA pairs and reasoning iteratively reviewed by a highly skilled clinical expert, our benchmark spans both image-level tasks (detection, localization, classification, diagnosis, captioning) and volume-level tasks following a hierarchical structure (detection $\rightarrow$ localization $\rightarrow$ classification $\rightarrow$ counting $\rightarrow$ captioning) that mirrors how a radiologist reconstructs a 3D understanding across slices. For localization, we specify both \textit{which frame} and \textit{where within the frame} each finding occurs, providing anatomical descriptions alongside bounding box coordinates.

Through our evaluation of 10 VLMs with metrics that separately measure textual accuracy, reasoning quality, and spatial grounding precision, we show that baseline VLMs demonstrate strong textual spatial understanding but struggle with pixel-level spatial grounding and accurate localization of findings. This reveals a gap between textual spatial understanding---recognizing anatomical regions described in language (e.g., the left or right hemisphere or the frontal lobe in brain MRI, and the medial compartment in knee MRI)---and pixel-level spatial reasoning, which requires precisely localizing findings in the image using spatial coordinates such as bounding boxes. This gap in spatial reasoning has not been captured by prior benchmarks, which either lack grounding evaluation or are limited to single-image settings. We address this limitation by fully fine-tuning Qwen3-VL-8B on our training set, which results in improved spatial grounding performance. This demonstrates that targeted supervision on spatially annotated medical data can help bridge this gap. To support future work, we open-source our benchmark, evaluation pipeline, and all model outputs at \url{https://lamawmouk.github.io/SGMRI-VQA} to encourage research on volumetric spatial grounding in knee and brain MRI data.

\vspace{0.3cm}
In summary, our contributions are as follows:
\begin{itemize}[leftmargin=*]
    \item We introduce \textbf{SGMRI-VQA}, a large-scale multi-frame VQA benchmark for knee and brain MRI images, with chain-of-thought reasoning and QA pairs iteratively reviewed by a highly skilled clinical expert. Image-level evaluation spans five tasks (detection, localization, classification, diagnosis, captioning), while volume-level evaluation follows a hierarchical structure (detection, localization, classification, counting, captioning) that mirrors how a radiologist reconstructs a 3D understanding across slices.

    \item We introduce cross-slice visual grounding with \textbf{spatially grounded localization}---providing both anatomical descriptions and bounding box coordinates with slice indices, jointly evaluating \textit{where} anatomically and \textit{where} spatially each finding occurs.

   \item We fine-tune \textbf{Qwen3-VL-8B} using full fine-tuning on our training set with bounding box supervision. Image-level QA pairs serve as fine-grained augmentation for volume-level training, providing slice-level supervision that helps the model learn spatial grounding within individual frames, and improves overall spatial precision.

    \item We compare the performance of our finetuned Qwen3-VL-8B against \textbf{10 VLMs} (3 proprietary models, 5 open-source models, and 2 domain-specific medical models), across two anatomical domains, multiple tasks, and at both image-level and volume-level granularity, quantifying the gap between textual competence and spatial grounding ability.
\end{itemize}

\begin{figure}[t]
\centering
\includegraphics[width=\textwidth]{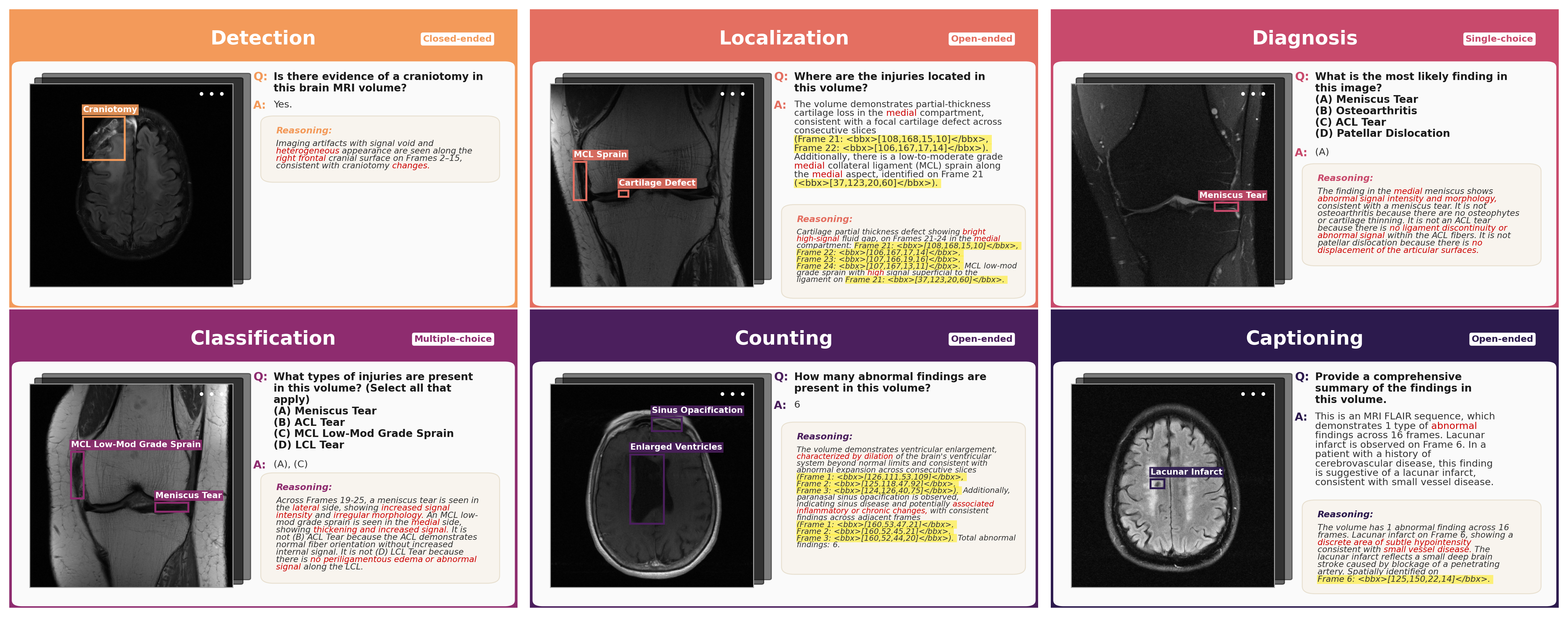}
\caption{\textbf{Examples of the diverse question tasks in SGMRI-VQA.} Each task is illustrated with real MRI slices and chain-of-thought reasoning, generated at both volume-level and image-level granularities across brain and knee domains, with tasks providing spatially grounded bounding box coordinates.}
\label{fig:task_examples}
\end{figure}

\section{Related Works}

\textbf{Medical Visual Question Answering.}\quad Medical VQA has progressed rapidly, from early benchmarks like VQA-RAD~\cite{lau2018dataset}, SLAKE~\cite{liu2021slake}, and PathVQA~\cite{he2020pathvqa} to large-scale evaluations spanning hundreds of thousands of questions across diverse modalities~\cite{zhang2023pmc,hu2024omnimedvqa,ye2024gmai,sun2024pathmmu}. Yet all of these benchmarks operate on single 2D images---none require models to reason across multiple frames or localize findings within a volumetric context. MedFrameQA~\cite{yu2025medframeqa} took a step toward multi-image evaluation with 2--5 YouTube video frames, but these are temporally related procedural images, not spatially contiguous slices from a single scan. Our SGMRI-VQA is the first benchmark that requires models to reason over a 12--60 sequential knee and brain MRI slices forming a continuous 3D representation, producing both chain-of-thought textual answers and spatially grounded bounding box coordinates with cross-slice indices.

\textbf{Vision-Language Models.}\quad Proprietary VLMs such as GPT-4o~\cite{achiam2023gpt}, Gemini~\cite{team2023gemini} achieve strong performance on general multimodal benchmarks, while open-source models---LLaVA~\cite{liu2023visual,munasinghe2025videoglamm}, InternVL~\cite{chen2024internvl}, Qwen-2VL/Qwen3-VL~\cite{bai2025qwen3,wang2024qwen2}, Eagle~2.5~\cite{chen2025eagle}---are closing this gap at the 7--8B parameter scale. In the medical domain, LLaVA-Med~\cite{li2023llava}, MiniGPT-Med~\cite{alkhaldi2024minigpt}, and MedMax~\cite{bansal2024medmax} have adapted these architectures through domain-specific pretraining or medical instruction tuning. However, both general and medical VLMs have been evaluated predominantly on single-image tasks. While architectures like Qwen3-VL~\cite{bai2025qwen3} support multi-image input, no existing benchmark tests whether these models can track and localize findings across volumetric medical data. Our SGMRI-VQA benchmark fills this gap by evaluating 10 of these models on multi-frame MRI reasoning with spatial grounding, chain-of-thought reasoning, and hierarchical task structure across two anatomical domains. Table~\ref{tab:benchmark_comparison} provides a comprehensive comparison of SGMRI-VQA with existing medical VQA benchmarks across key evaluation dimensions.

\textbf{Visual Grounding.}\quad Visual grounding~\cite{xiao2025towards, xue2025point, li2022grounded, liu2024grounding} in natural images has advanced from specialized models like GLIP~\cite{li2022grounded} and Grounding DINO~\cite{liu2024grounding} to foundation models like SAM~\cite{ravi2024sam2segmentimages}, with medical extensions such MedSAM2~\cite{MedSAM2}. In medical VQA, grounded benchmarks remain limited to single 2D images: GEMeX~\cite{liu2025gemex} provides 1.6M grounded chest X-ray questions, VinDr-CXR-VQA~\cite{nguyen2025vindr} offers 17.6K grounded questions, and NOVA~\cite{bercea2025nova} evaluates anomaly localization on brain MRI across 906 scans. NOVA is the closest prior work, but evaluates only single 2D slices. None of these benchmarks require cross-slice grounding---specifying both which slice and where within it a finding occurs. In contrast, we are the first to evaluate knee and brain MRIs for textual spatial understanding (e.g., the left or right hemisphere or the frontal lobe in brain MRI, and the medial compartment in knee MRI) alongside pixel-level spatial reasoning using bounding box coordinates with cross-slice indices.

\textbf{Volumetric Image Understanding.}\quad Volumetric understanding has been studied through task-specific architectures like UNETR~\cite{hatamizadeh2022unetr}, 3D U-Net~\cite{cciccek20163d}, and SegFormer3D-Net~\cite{perera2024segformer3d} process volumes with specialized 3D architectures. Conversely, our SGMRI-VQA benchmark evaluates standard multi-image VLMs that receive knee and brain MRI slices as a sequence of 2D frames and reason about their spatial relationships.

\begin{table}[t]
\centering
\caption{\textbf{Comparison with medical VQA benchmarks.} O = Open-ended, C = Closed (Yes/No), SC = Single-choice, MC = Multiple-choice, CoT = chain-of-thought reasoning.}
\label{tab:benchmark_comparison}
\setlength{\tabcolsep}{3pt}
\resizebox{\textwidth}{!}{%
\begin{tabular}{lccccccc}
\toprule
\rowcolor{gray!15}
\textbf{Dataset} & \textbf{Domain} & \textbf{Images} & \textbf{QA Pairs} & \textbf{Types} & \textbf{Grounded} & \textbf{Multi-Img} & \textbf{Volume} \\
\midrule
VQA-RAD~\cite{lau2018dataset}        & Radiology   & 315    & 3.5K  & O/C           & \budui & \budui & \budui \\
SLAKE~\cite{liu2021slake}            & Radiology   & 642    & 14K   & O/C           & \budui & \budui & \budui \\
PathVQA~\cite{he2020pathvqa}         & Pathology   & 4.3K   & 32K   & O/C           & \budui & \budui & \budui \\
PMC-VQA~\cite{zhang2023pmc}          & Multi-modal & 149K   & 227K  & Open          & \budui & \budui & \budui \\
OmniMedVQA~\cite{hu2024omnimedvqa}   & 12 mod.     & 118K   & 128K  & Closed        & \budui & \budui & \budui \\
GMAI-MMBench~\cite{ye2024gmai}       & 38 mod.     & 284 sets & 26K & Closed        & \budui & \budui & \budui \\
PathMMU~\cite{sun2024pathmmu}        & Pathology   & 24K    & 33K   & MC            & \budui & \budui & \budui \\
GEMeX~\cite{liu2025gemex}            & CXR         & 151K   & 1.6M  & O/C/SC/MC     & \dui   & \budui & \budui \\
ReXVQA~\cite{pal2025rexvqa}          & CXR         & 160K   & 696K  & Closed        & \budui & \budui & \budui \\
MedFrameQA~\cite{yu2025medframeqa}   & Multi-modal & 9K     & 2.9K  & Closed        & \budui & \dui~(2--5) & \budui \\
NOVA~\cite{bercea2025nova}           & Brain MRI   & 906    & 2.7K  & Open          & \dui   & \budui & \budui \\
VinDr-CXR-VQA~\cite{nguyen2025vindr} & CXR        & 4.4K   & 17.6K & O/C           & \dui   & \budui & \budui \\
\midrule
\rowcolor{lightblue}
\textbf{SGMRI-VQA (Ours)} & \textbf{Brain + Knee} & \textbf{50K+} & \textbf{41K} & \textbf{O/C/SC/MC/CoT} & \textbf{\dui} & \textbf{\dui~(over a dozen)} & \textbf{\dui} \\
\bottomrule
\end{tabular}
}
\end{table}

\section{SGMRI-VQA Benchmark}

\subsection{Data Sources}

We build on the fastMRI+ dataset~\cite{zhao2022fastmri+}, which provides expert radiologist annotations with pixel-level bounding boxes across two anatomical domains. The brain subset contains axial FLAIR and T1-weighted volumes spanning nine finding categories, while the knee subset contains sagittal proton-density weighted sequences with fat saturation spanning 22 categories across ligamentous, meniscal, cartilage, and osseous pathology. Each annotation provides bounding box coordinates in $\{x, y, \text{width}, \text{height}\}$ format with an associated finding label, localized to the specific slices where each abnormality is visible.

\subsubsection{Data Preprocessing}

We process raw HDF5 volumes to extract 2D slices as grayscale PNG images, applying orientation correction and min-max normalization. Bounding box annotations from the fastMRI+ CSV files are matched to individual slices using the volume identifier and slice index, while volumes without valid annotations are filtered out. After filtering, 1,970 volumes are retained (996 brain, 974 knee), comprising 50,672 slices, of which 13,202 contain at least one bounding box, totaling 23,724 bounding boxes across both anatomical domains.

\subsection{QA Generation Pipeline}

We use GPT-4o to generate visually grounded question-answer pairs with chain-of-thought reasoning at two granularities: image-level (single slice) and volume-level (full 3D scan). Prompt design is guided by clinical expert input which defines task-specific rules, output format examples, radiological conventions, and reasoning templates for each task across both anatomical domains. For each sample, GPT-4o is prompted as a domain-specific expert (neuroradiologist for brain, musculoskeletal radiologist for knee) and receives: (1) the original MRI image, (2) the same image with numbered bounding box overlays as visual guides to finding locations, (3) bounding box coordinates and labels as structured text, and (4) clinical context including modality, diagnosis, and slice-level labels. The prompt enforces radiological conventions: for brain MRI, the left side of the image corresponds to the right side of the patient's brain; for knee MRI, medial and lateral orientation is determined from tibial plateau morphology and the fibula position, since sagittal slices lack explicit laterality cues. All reasoning must be grounded in visual features observed in the original clean MRI image---signal intensity, shape, borders, and texture---rather than inferred from the bounding box overlays, which serve only as spatial guides. GPT-4o is instructed to reference findings by actual names and anatomical locations rather than region numbers, distinguish co-occurring findings of the same type by anatomy or frame number, and include \texttt{<bbx>[x, y, w, h]</bbx>} coordinates to anchor observations to spatial locations. For each task, the physician defines a structured JSON output template with fields for task type, question, answer, and reasoning, with example outputs that establish the expected clinical terminology and level of detail. All questions are constrained to require viewing the specific image to answer, ensuring the benchmark evaluates genuine visual understanding rather than text-based knowledge recall.

Figure~\ref{fig:prompt_cards} (Appendix) provides an overview of the four prompt templates, and Figure~\ref{fig:task_examples} illustrates representative examples from each task category.

\subsubsection{Image-Level Questions}

For each MRI slice, GPT-4o generates five Q\&A pairs across four question formats (open-ended, closed-ended, single-choice, multiple-choice), each assigned to a different task: \textit{detection} (identify whether a specific finding is present), \textit{localization} (describe the spatial location anatomically and provide bounding box coordinates in \texttt{<bbx>[x, y, w, h]</bbx>} format), \textit{classification} (classify the abnormality from predefined choices), \textit{diagnosis} (identify the most likely condition), and \textit{captioning} (provide a clinical description of the image findings). Together, these five tasks evaluate the core competencies required for complex MRI interpretation---from recognizing abnormalities to precisely localizing and characterizing them---alongside the chain-of-thought reasoning essential for clinical decision-making. For localization, we provide anatomical descriptions (e.g., ``in the right hemisphere,'' ``in the medial compartment'') and pixel-level bounding box coordinates, enabling evaluation of both reasoning quality and spatial accuracy. We use single-frame QA as auxiliary supervision when fine-tuning on multi-frame volumetric data to improve the model's understanding of spatial contiguity across slices.

\subsubsection{Volume-Level Questions}

For volume-level QA, GPT-4o receives all slices from a volume as sequential frames and generates five Q\&A pairs following a hierarchical diagnostic progression: \textit{detection} (identify finding categories present, referencing specific frames), \textit{localization} (describe spatial locations with anatomical descriptions and \texttt{<bbx>} coordinates with frame references), \textit{counting} (quantify lesion burden across the volume), \textit{classification} (classify findings based on signal characteristics and morphology), and \textit{captioning} (generate a comprehensive clinical summary with structured reasoning). This hierarchy evaluates the same core interpretation capabilities at the volumetric scale and mirrors how a radiologist reconstructs a 3D understanding across slices---each question builds upon the preceding ones.

\begin{figure}[t]
\centering
\includegraphics[width=\linewidth]{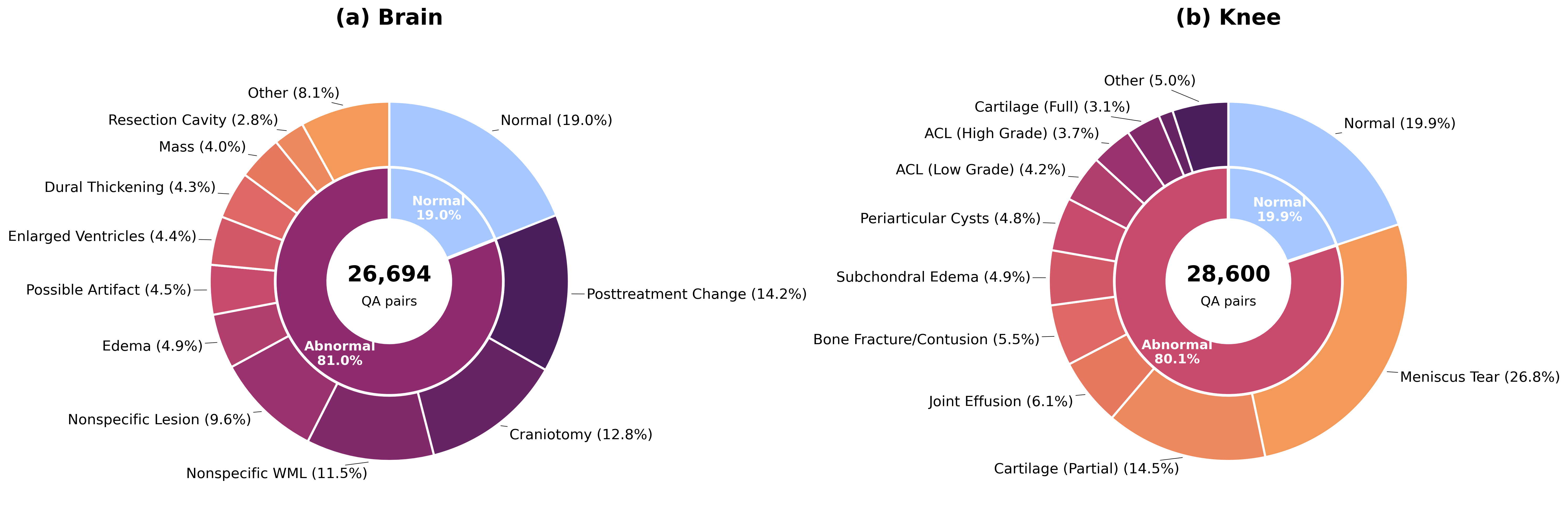}
\caption{\textbf{Finding category distribution.} Inner ring: normal vs.\ abnormal. Outer ring: finding categories proportional to QA pair count.}
\label{fig:finding_distribution}
\end{figure}

\begin{figure}[t]
\centering
\includegraphics[width=\linewidth]{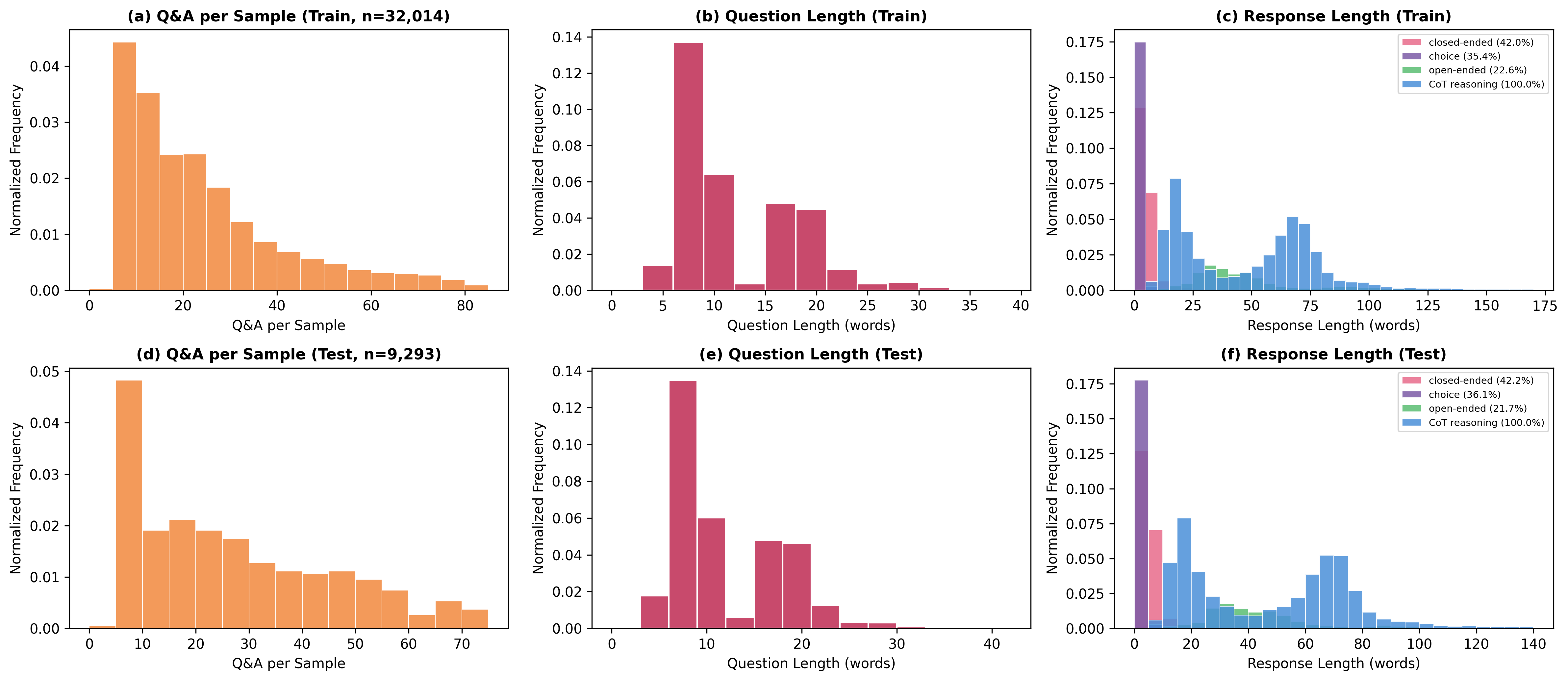}
\caption{\textbf{Distribution statistics of the SGMRI-VQA dataset.} Top row (a--c): training set (32,014 QA pairs). Bottom row (d--f): test set (9,293 QA pairs). (a,d) Distribution of Q\&A pairs per sample. (b,e) Question length distribution. (c,f) Response length distribution stratified by answer format---closed-ended, choice, and open-ended---and chain-of-thought (CoT) reasoning traces. Normalized frequency is the item count divided by the total count across all items.}
\label{fig:dataset-stats}
\end{figure}

\subsection{Quality Assurance}

GPT-4o initially generates 41,649 QA pairs across both domains. While the raw outputs are clinically reasonable, systematic quality assurance is required to ensure spatial consistency, anatomical correctness, and reasoning completeness. Clinical expertise was used as input to define task-specific quality criteria for each cleaning stage, specifying acceptable anatomical terminology, laterality conventions, and reasoning standards per domain and task type.

Despite producing clinically reasonable text, GPT-4o exhibits systematic spatial reasoning failures that require correction (Figure~\ref{fig:gpt4o_errors}). For brain MRI, GPT-4o frequently assigns incorrect hemisphere laterality---stating ``right frontal lobe'' when the bounding box coordinates place the finding in the left hemisphere, or vice versa---affecting approximately 18\% of localization and captioning entries. The model also generates imprecise spatial references (e.g., ``periventricular region'', ``frontal lobe'') instead of standardized hemisphere labels. Thus, we normalize all spatial references to left hemisphere, right hemisphere, or bilateral hemispheres using the bounding box x-coordinate under radiological convention, with enlarged ventricle findings crossing the midline labeled as bilateral. For knee MRI, the physician identified that GPT-4o consistently misassigns medial/lateral labels in sagittal knee images, a systematic error affecting over 1,500 entries. To address this, the physician manually annotated the fibula side for each volume through a purpose-built annotation interface (Figure~\ref{fig:fibula_annotator}, Appendix), since the fibula is always lateral to the knee joint. All laterality labels were then corrected using each finding's bounding box position relative to the image midline and the annotated fibula side, resolving the majority of localization errors across both image-level and volume-level QA pairs. Additionally, GPT-4o frequently omits co-occurring findings in detection---for volumes with multiple diagnoses, the model typically describes only the most prominent finding while omitting others present in the ground truth. Thus, erroneous frame references and trailing incomplete sentences are corrected.

\subsubsection{Domain Expert Review}

Clinical expertise was integrated throughout the quality control process---defining the acceptance criteria for each task type, specifying what constitutes clinically accurate reasoning, and establishing domain-specific standards for anatomical descriptions, laterality, and diagnostic terminology. Using a purpose-built web-based interface with annotated MRI images, generated QA pairs, ground truth annotations, and an animated viewer for frame-by-frame volume inspection, the physician reviews all QA pairs (Figure~\ref{fig:web_annotator}, Appendix). GPT-4o occasionally omits findings present in the ground truth or produces incomplete reasoning. To address this, the physician appends modality-specific descriptions for missing findings, validates mentioned findings against ground truth annotations to remove hallucinations, and replaces tautological or clinically imprecise descriptions with meaningful alternatives. The testing set receives full coverage (all 222 brain and 155 knee testing volumes reviewed), while for the training set the physician reviews over 300 volumes across both domains in an iterative process---verifying correctness and validity of the automated corrections and providing feedback that informs subsequent cleaning iterations. The final dataset comprises 41,307 QA pairs (32,142 image-level and 9,165 volume-level) split into 32,014 training and 9,293 testing pairs, each with a question, answer, reasoning trace, and ground truth annotations with bounding box coordinates.
Table~\ref{tab:dataset_stats} summarizes the statistics of the SGMRI-VQA dataset and Figure~\ref{fig:dataset-stats} visualizes the distribution characteristics.

\begin{table}[t]
\centering
\caption{\textbf{SGMRI-VQA dataset statistics.} Each QA pair includes a question, answer, reasoning trace, and ground-truth annotations with bounding box coordinates.}
\label{tab:dataset_stats}
\setlength{\tabcolsep}{4pt}
\begin{tabular}{lcccc}
\toprule
\rowcolor{gray!15}
\textbf{Domain} & \textbf{Split} & \textbf{Image-Level} & \textbf{Volume-Level} & \textbf{Total} \\
\midrule
\multirow{2}{*}{Brain}
& Train & 12,193 & 3,231 & 15,424 \\
& Test  & 3,704  & 1,064 & 4,768 \\
\midrule
\multirow{2}{*}{Knee}
& Train & 12,495 & 4,095 & 16,590 \\
& Test  & 3,750  & 775   & 4,525 \\
\midrule
\rowcolor{lightblue}
\textbf{Total} &  & \textbf{32,142} & \textbf{9,165} & \textbf{41,307} \\
\bottomrule
\end{tabular}
\end{table}

The distribution of finding categories across both domains is shown in Figure~\ref{fig:finding_distribution}. The dataset is disease-dense: the majority of QA pairs target abnormal findings, reflecting clinical practice where MRI scans are typically ordered when an abnormality is suspected and detecting and localizing real findings across diverse categories is more challenging than normal scans.

\section{Experiments}

\subsection{Experimental Setup}

\textbf{Datasets.}\quad We evaluated two anatomical domains drawn from the fastMRI+ dataset: brain MRI with 670 training and 222 validation volumes and knee MRI with 819 training and 155 validation volumes. The final benchmark comprises 41,307 QA pairs. Evaluation is conducted at two granularities: volume-level, where models process the full 3D scan as a multi-frame sequence, and image-level, where models analyze a single MRI slice.

\textbf{Candidate models.}\quad We evaluated 10 VLMs on SGMRI-VQA. For proprietary models, we test three models: GPT-4o~\cite{achiam2023gpt}, Gemini 2.5 Pro~\cite{team2023gemini}, and Gemini 2.5 Flash~\cite{team2023gemini}. For open-source models, we evaluate five models at the 7--8B parameter scale: Qwen2.5-VL-7B~\cite{wang2024qwen2}, Qwen3-VL-8B~\cite{bai2025qwen3}, Eagle 2.5-8B~\cite{chen2025eagle}, InternVL2.5-8B~\cite{chen2024internvl}, and LLaVA-Video-7B~\cite{zhang2024llava}. We additionally evaluate two domain-specific medical VLMs at the image level only: LLaVA-Med-v1.5~\cite{li2023llava} and MedGemma-1.5 (4B)~\cite{sellergren2025medgemma}. To enforce determinism in open-source models, we set \texttt{do\_sample=False} and \texttt{temperature=0}. For proprietary models, we use \texttt{temperature=0}. All selected models support the concurrent image frames required for volume-level evaluation, so no slices were subsampled. All models are evaluated under identical zero-shot conditions.

\textbf{Fine-tuning Qwen3-VL.}\quad We fine-tune Qwen3-VL-Instruct-8B on the 32,014-sample training split containing both image and video data, using 8 H100 GPUs with a global batch size of 64 (per-device batch size 1, gradient accumulation 8), cosine learning rate schedule with 3\% warmup, learning rate $2 \times 10^{-5}$, and bf16 mixed precision over 2 epochs. We perform full SFT using DeepSpeed ZeRO Stage~3, where the vision encoder and multi-modal projector are frozen and only the language model backbone is trained.

\textbf{Evaluation Metrics.}\quad We use three complementary evaluation metrics. \textit{A-Score} measures factual answer accuracy for detection, counting, classification, and diagnosis tasks. Scoring differs by question format: closed-ended questions use exact match on Yes/No, single-choice uses exact match on the selected option letter, multiple-choice uses F1 over the selected option set, and open-ended uses the average of keyword recall and semantic similarity via SentenceTransformer embeddings. \textit{AR-Score} evaluates free-text clinical reasoning quality for captioning and localization tasks as a weighted combination of GPT-4o-mini judge scoring (weight 0.4), BERTScore~\cite{zhang2020bertscoreevaluatingtextgeneration} F1 (0.2), smoothed BLEU (0.2), and ROUGE-L F1 (0.2). For localization tasks, bounding box coordinates and frame references are stripped from both prediction and reference before scoring, so the judge evaluates anatomical description quality rather than numeric coordinates. \textit{V-Score} measures pixel-level spatial grounding accuracy as mean IoU between predicted and ground-truth bounding boxes for localization tasks, complementing the textual anatomical descriptions evaluated by AR-Score---together, these two metrics jointly assess \textit{where} anatomically (in text) and \textit{where} spatially (in pixel coordinates) each finding occurs. Predicted bounding boxes are extracted via a multi-format parser, matched to ground truth using frame-aware IoU matrices and the Hungarian algorithm, and the V-Score is the mean IoU over all ground-truth boxes.

\subsection{Qualitative Results}

\begin{figure}[t]
\centering
\includegraphics[width=\textwidth]{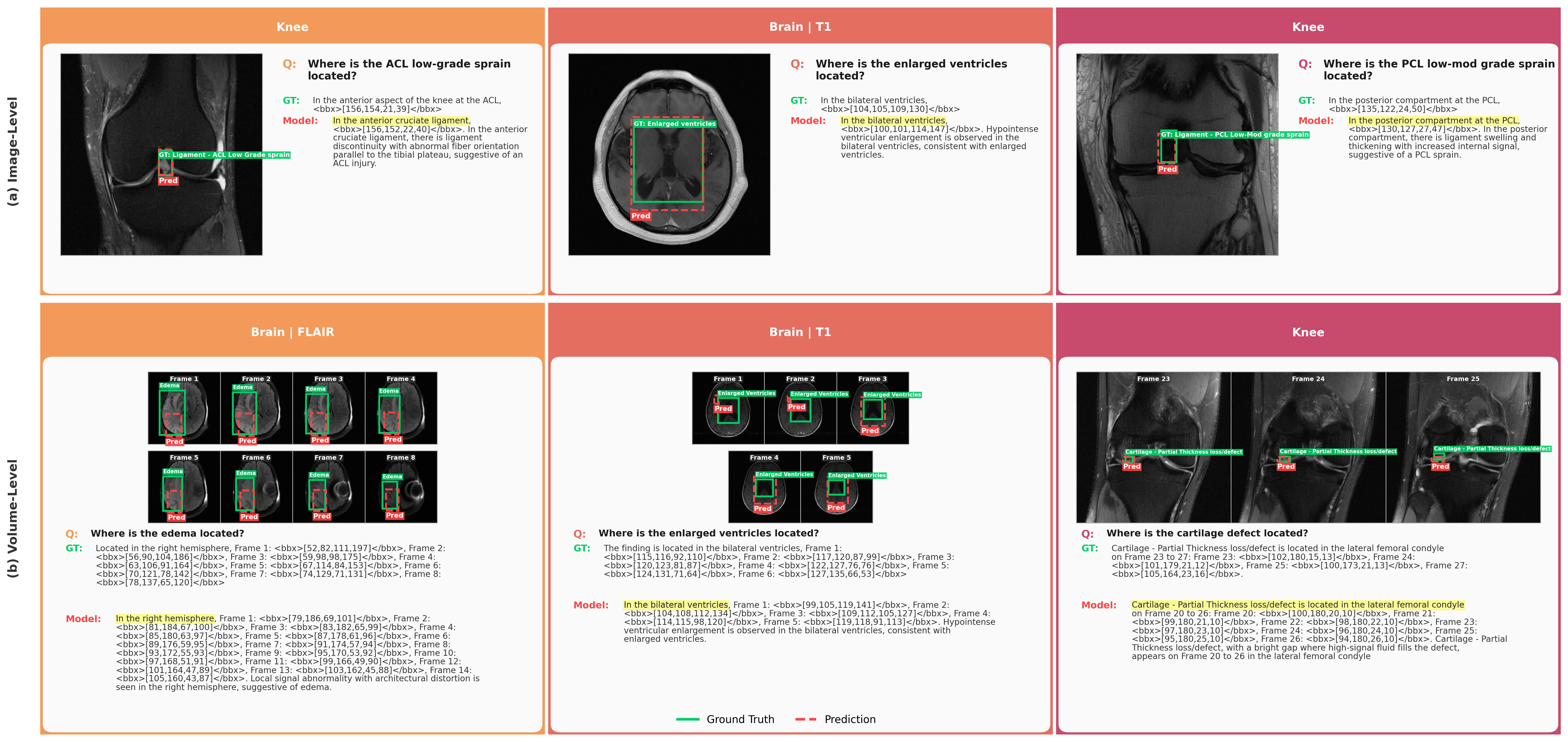}
\caption{Qualitative localization examples from the finetuned Qwen3-VL-8B. \textbf{(a)} Image-level predictions on single MRI slices. \textbf{(b)} Volume-level predictions across multi-frame sequences. Green boxes denote ground-truth annotations with pathology labels; red dashed boxes denote model predictions. For both image-level and volume-level predictions, we show the input image(s), evaluation question, ground-truth answer, and model output.}
\label{fig:combined-bbox}
\end{figure}

To illustrate the spatial grounding capabilities of our fine-tuned model, we present qualitative examples of the finetuned Qwen3-VL-8B predictions at both evaluation granularities. Figure~\ref{fig:combined-bbox} shows representative localization predictions alongside ground-truth bounding boxes. The top row displays image-level examples, where models analyze a single MRI slice and produce bounding boxes for all visible findings. The bottom row shows volume-level examples, where models must additionally identify the correct frames containing pathological findings across a multi-frame sequence. The examples span both brain and knee MRI, demonstrating the model's ability to identify the anatomical regions in image-level and produce spatially overlapping bounding boxes with ground-truth annotations.

\subsection{Quantitative Results}

\begin{table}[t]
\centering
\caption{Image-level task performance on the MRI-VQA test set. Metrics: A-Score (exact-match accuracy) for closed-ended tasks, V-Score (mIoU) for spatial grounding, AR-Score (GPT-4o-mini judge + NLG) for open-ended generation. All values in \%. Best in \textbf{bold}, second best \underline{underlined}.}
\label{tab:image-taskwise}
\setlength{\tabcolsep}{4pt}
\resizebox{\textwidth}{!}{
\begin{tabular}{lccccccc}
\toprule
\rowcolor{gray!15}
& \textbf{Detection} & \textbf{Classification} & \textbf{Diagnosis} & \multicolumn{2}{c}{\textbf{Localization}} & \textbf{Captioning} & \\
\rowcolor{gray!15}
\textbf{Model} & A-Score & A-Score & A-Score & V-Score & AR-Score & AR-Score & \textbf{Avg.} \\
\midrule
\multicolumn{8}{l}{\textit{Proprietary Models}} \\
GPT-4o                  & 50.74          & 52.49          & 44.71          & 3.41          & 26.22         & 16.90         & 32.41 \\
Gemini-2.5-Pro          & 19.24          & 31.33          & 34.32          & 3.33          & 14.89         & 11.91         & 19.17 \\
Gemini-2.5-Flash        & \underline{67.36} & \underline{86.31} & \underline{89.68} & \underline{5.95} & \underline{27.85} & \underline{21.72} & \underline{49.81} \\
\midrule
\multicolumn{8}{l}{\textit{Open-Source Models (7--8B)}} \\
LLaVA-Video-7B          & 52.41          & 65.41          & 71.58          & 2.29          & 23.11         & 13.31         & 38.02 \\
Eagle2.5-8B             & 26.94          & 57.55          & 61.39          & 2.59          & 23.98         & 15.41         & 31.31 \\
Qwen3-VL-8B             & 46.11          & 76.02          & 80.29          & 3.11          & 24.55         & 19.86         & 41.66 \\
InternVL2.5-8B          & 22.25          & 37.33          & 30.43          & 1.82          & 25.78         & 14.37         & 22.00 \\
Qwen2.5-VL-7B           & 5.29           & 50.68          & 39.88          & 1.51          & 26.75         & 18.19         & 23.72 \\
\midrule
\multicolumn{8}{l}{\textit{Domain-Specific Medical VLMs (image-level only)}} \\
LLaVA-Med-v1.5 (7B)    & 6.90           & 85.36          & 59.18          & 0.00          & 23.43         & 12.20         & 31.18 \\
MedGemma-1.5 (4B)      & 20.11          & 65.98          & 64.61          & 2.77          & 24.93         & 17.24         & 32.61 \\
\midrule
\rowcolor{lightblue}
\textbf{Ours (Qwen3-VL-8B-SGMRIVQA)} & \textbf{95.11} & \textbf{97.56} & \textbf{94.50} & \textbf{15.51} & \textbf{28.99} & \textbf{25.05} & \textbf{59.45} \\
\bottomrule
\end{tabular}
}
\end{table}

\begin{table}[t]
\centering
\caption{Volume-level task performance on the MRI-VQA test set. All values in \%. Best in \textbf{bold}, second best \underline{underlined}.}
\label{tab:volume-taskwise}
\setlength{\tabcolsep}{4pt}
\resizebox{\textwidth}{!}{
\begin{tabular}{lcccccccc}
\toprule
\rowcolor{gray!15}
& \textbf{Detection} & \textbf{Counting} & \textbf{Classification} & \multicolumn{2}{c}{\textbf{Localization}} & \textbf{Captioning} & \\
\rowcolor{gray!15}
\textbf{Model} & A-Score & A-Score & A-Score & V-Score & AR-Score & AR-Score & \textbf{Avg.} \\
\midrule
\multicolumn{8}{l}{\textit{Proprietary Models}} \\
GPT-4o                  & 70.57          & \underline{35.97} & 67.51          & 1.20          & \underline{24.83} & 20.19         & \underline{36.71} \\
Gemini-2.5-Pro          & 16.03          & 4.08           & 28.21          & 0.26          & 12.42         & 12.71         & 12.29 \\
Gemini-2.5-Flash        & 54.89          & 17.66          & 56.31          & \underline{1.83} & 23.27         & \underline{22.27} & 29.37 \\
\midrule
\multicolumn{8}{l}{\textit{Open-Source Models (7--8B)}} \\
Qwen3-VL-8B             & 62.23          & 30.98          & 64.72          & 0.16          & 21.93         & 19.05         & 33.18 \\
Eagle2.5-8B             & \underline{70.65} & 16.30          & \underline{70.73} & 0.02          & 19.07         & 16.20         & 32.16 \\
InternVL2.5-8B          & 28.80          & 14.95          & 44.01          & 0.41          & 17.66         & 16.29         & 20.35 \\
LLaVA-Video-7B          & 37.23          & 13.59          & 48.94          & 0.06          & 16.26         & 12.36         & 21.41 \\
Qwen2.5-VL-7B           & 24.18          & 25.27          & 62.54          & 0.07          & 15.63         & 16.67         & 24.06 \\
\midrule
\rowcolor{lightblue}
\textbf{Ours (Qwen3-VL-8B-SGMRIVQA)} & \textbf{99.18} & \textbf{37.77} & \textbf{97.70} & \textbf{5.97} & \textbf{28.24} & \textbf{26.54} & \textbf{49.23} \\
\bottomrule
\end{tabular}
}
\end{table}

\subsubsection{Image-Level Task Performance}

As shown in Table~\ref{tab:image-taskwise}, our fine-tuned model (Qwen3-VL-8B-SGMRIVQA) achieves the highest overall average score of 59.45\%, outperforming all proprietary and open-source baselines. On Detection, our model leads with 95.11\%, far ahead of Gemini-2.5-Flash (67.36\%) and LLaVA-Video-7B (52.41\%). On Classification and Diagnosis, it leads all models with scores of 97.56\% and 94.50\% respectively, surpassing Gemini-2.5-Flash (86.31\% and 89.68\%) and purpose-built medical models like LLaVA-Med-v1.5 (85.36\% and 59.18\%). Localization proves to be the most differentiating task: our model achieves a V-Score of 15.51---more than 2.6$\times$ Gemini-2.5-Flash's 5.95---and an AR-Score of 28.99, compared to 27.85 for Gemini-2.5-Flash. On Captioning, our model leads with 25.05, outperforming Gemini-2.5-Flash (21.72) and Qwen3-VL-8B (19.86). These results demonstrate that domain-specific fine-tuning of an 8B open-source model consistently matches or exceeds larger proprietary models across nearly all medical imaging tasks.

\subsubsection{Volume-Level Task Performance}

Table~\ref{tab:volume-taskwise} reports volumetric performance across detection, counting, classification, localization, and captioning tasks. While several baseline models achieve moderate performance on detection and classification (e.g., 70.65\% detection for Eagle2.5-8B and 70.73\% classification for Eagle2.5-8B), spatial grounding remains extremely challenging, with most models achieving near-zero V-Scores for localization. Our fine-tuned Qwen3-VL-8B-SGMRIVQA substantially improves volumetric reasoning and grounding, achieving the best detection (99.18\%), counting (37.77\%), classification (97.70\%), localization V-Score (5.97\%), localization AR-Score (28.24\%), and captioning performance (26.54\%). Overall, our model achieves the highest average score of 49.23\%, significantly outperforming both proprietary and open-source baselines.

\textbf{Fine-tuning.}\quad Across both image-level (Table~\ref{tab:image-taskwise}) and volume-level (Table~\ref{tab:volume-taskwise}) evaluations, fine-tuning substantially improves performance, particularly for spatial grounding and volumetric reasoning. Our fine-tuned model, Qwen3-VL-8B-SGMRIVQA, achieves the best overall performance on both settings, reaching an average score of 59.45 at the image level and 49.23 at the volume level. The largest gains are observed in spatial grounding tasks: at the image level, localization improves to 15.51 V-Score and 28.99 AR-Score, significantly outperforming the best baseline (5.95 V-Score from Gemini-2.5-Flash). At the volume level, our model achieves the strongest spatial grounding performance (5.97 V-Score and 28.24 AR-Score), while most baselines obtain near-zero localization scores. These results show that targeted fine-tuning on spatially annotated MRI data substantially improves both grounding and reasoning capabilities.

\textbf{Closed-source vs.\ open-source.}\quad Closed-source models (GPT-4o, Gemini-2.5-Pro, and Gemini-2.5-Flash) demonstrate competitive performance on certain tasks. Gemini-2.5-Flash achieves strong image-level classification (86.31\%) and diagnosis (89.68\%), while GPT-4o achieves reasonable volume-level detection (70.57\%) and classification (67.51\%). However, their performance on spatial grounding remains limited, with all proprietary models achieving low localization V-Scores. In contrast, open-source models show stronger detection and counting capabilities and provide a more flexible foundation for domain-specific adaptation through fine-tuning. Our results demonstrate that a fine-tuned open-source model can outperform both proprietary and open-source baselines across most tasks, highlighting the importance of domain-specific training for volumetric medical reasoning.

\section{Conclusion}

We introduce SGMRI-VQA, the first large-scale multi-frame VQA benchmark for spatially grounded reasoning across volumetric MRI. Our benchmark integrates slice-indexed visual grounding, hierarchical task evaluation, and domain-expert-aligned chain-of-thought reasoning across 41,307 QA pairs spanning both knee and brain 3D MRI images. Evaluation of 10 VLMs reveals a critical gap: while models often demonstrate clinically reasonable textual spatial understanding---correctly identifying anatomical regions described in language, such as the left or right hemisphere or the frontal lobe in brain MRI, or the medial compartment in knee MRI---they struggle with pixel-level spatial grounding, limiting their ability to accurately localize findings using bounding box coordinates and cross-slice indices.
Fine-tuning Qwen3-VL-8B on our training set improves grounding performance, demonstrating that targeted supervision could bridge this gap, yet a lot more data is needed in this domain for substantial improvements.

\textbf{Limitations.}\quad Although QA pairs were iteratively reviewed with clinical expertise, very small annotation noise may persist in reasoning traces. The benchmark currently spans two anatomical domains; extending to additional modalities and imaging protocols would further validate generalizability. We evaluated open-source models on the 7--8B scale; larger models (70B+) may exhibit different grounding characteristics but were excluded due to computational constraints.

\textbf{Future Directions.}\quad Beyond benchmarking, SGMRI-VQA opens several impactful research directions. The radiologist-aligned reasoning traces paired with spatially grounded coordinates provide a natural foundation for reinforcement learning approaches such as GRPO, where a reward function over correctly localized findings and accurate diagnoses could jointly optimize reasoning quality and spatial precision---without requiring additional annotations. More broadly, the hierarchical task structure and built-in clinical reasoning make SGMRI-VQA a stepping stone toward VLMs that function as clinical co-pilots: models that not only detect and classify findings but transparently explain their spatial reasoning, building clinician trust through explainable, grounded decision pathways.

\begin{ack}
We thank the Georgia Institute of Technology Partnership for an Advanced Computing Environment (PACE) and Lambda's Research Grant Program for AI for providing the computational resources that supported this work.
\end{ack}

\bibliographystyle{plainnat}
\bibliography{main}

\newpage
\appendix

\section{Appendix}

\subsection{Quality Assurance Interface}
\label{sec:qa_interface}

\begin{figure}[h!]
\centering
\includegraphics[width=\textwidth]{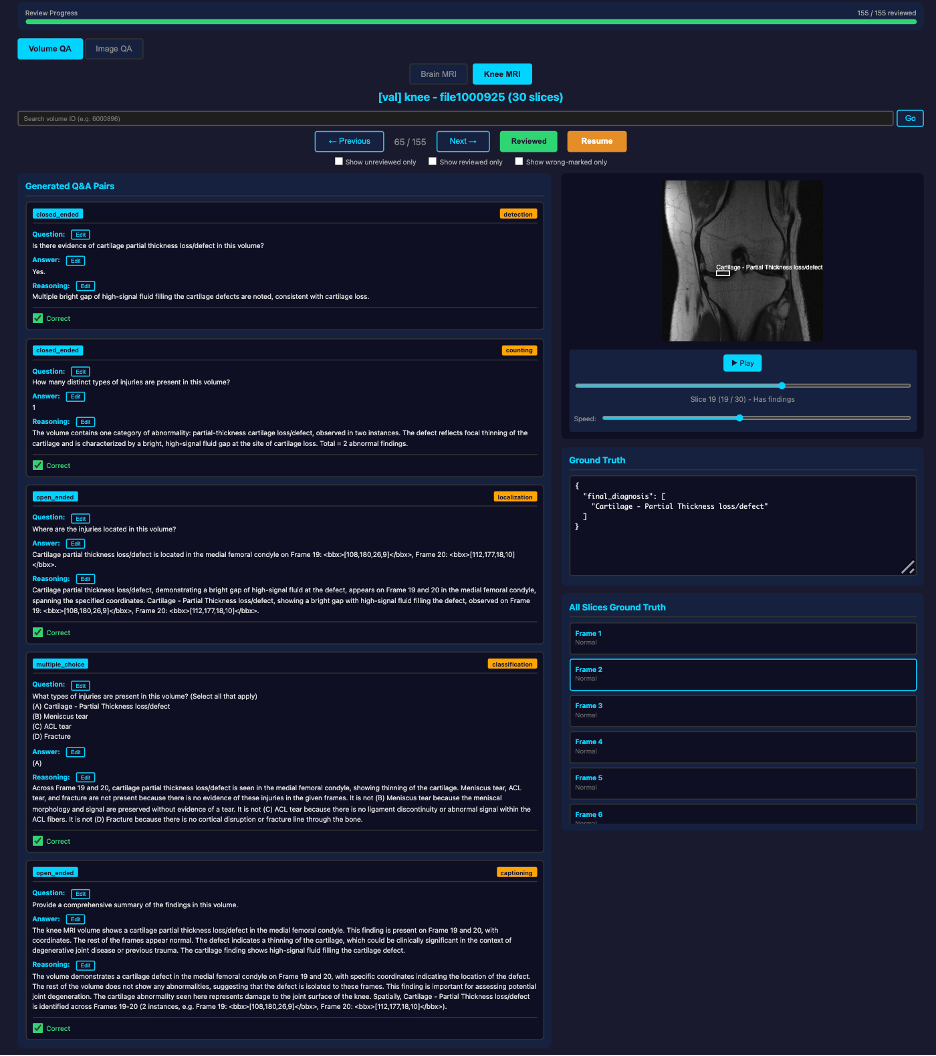}
\caption{\textbf{Web-based quality assurance interface.} The purpose-built annotation tool used by the physician for reviewing and correcting GPT-4o-generated QA pairs. The interface displays annotated MRI images with bounding box overlays, the generated question-answer pairs with reasoning traces, ground truth annotations, and an animated viewer for frame-by-frame volume inspection.}
\label{fig:web_annotator}
\end{figure}

\begin{figure}[h!]
\centering
\includegraphics[width=\textwidth]{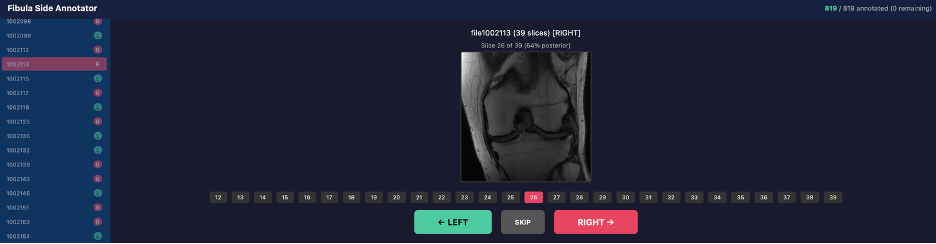}
\caption{\textbf{Fibula side annotation interface.} A purpose-built web tool for annotating fibula laterality across knee MRI volumes. Since the fibula is always lateral to the knee joint, annotating which side of the image the fibula appears on determines the medial/lateral orientation for each volume.}
\label{fig:fibula_annotator}
\end{figure}

\begin{figure}[h!]
\centering
\includegraphics[width=\textwidth]{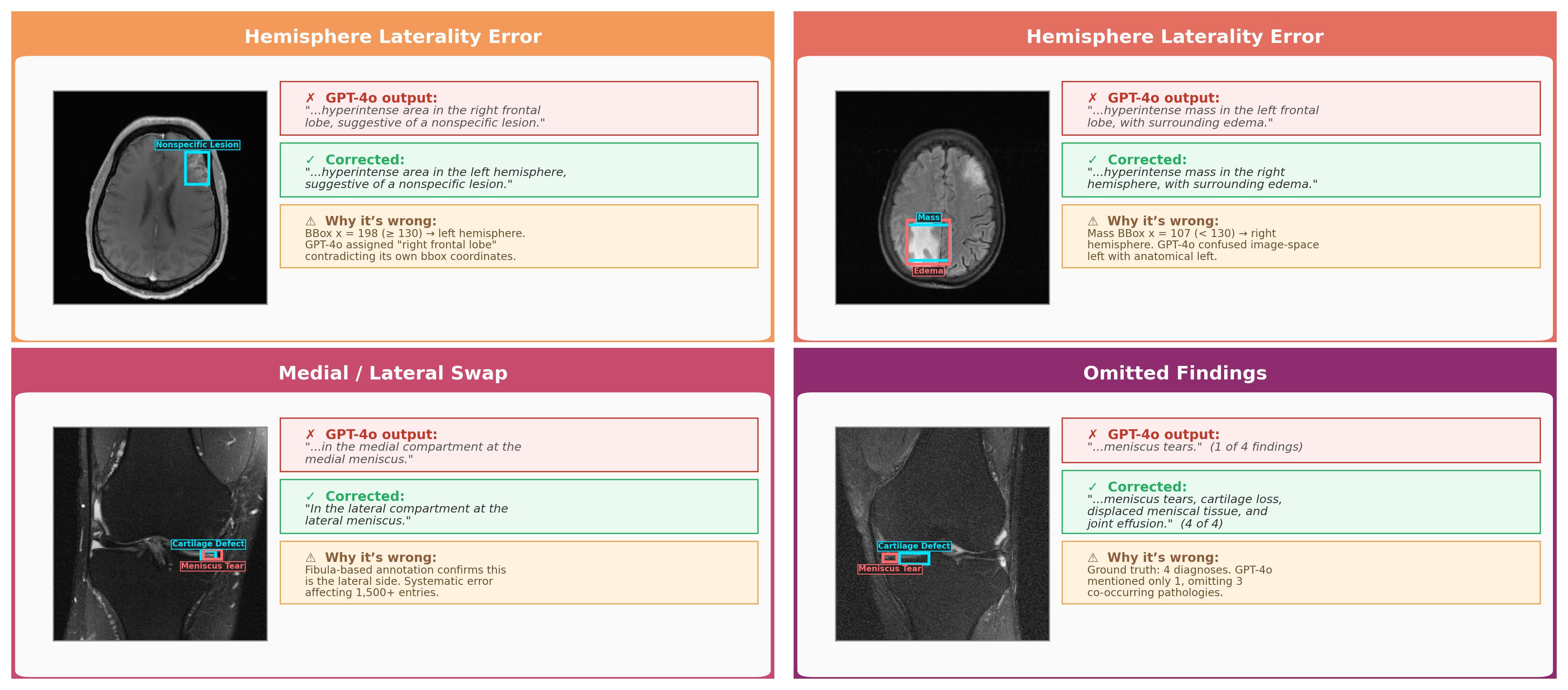}
\caption{\textbf{Examples of GPT-4o hallucination errors corrected during quality assurance.} Hemisphere laterality spatial errors: GPT-4o assigns incorrect hemisphere labels that contradict its own bounding box coordinates. Knee medial/lateral spatial swap and omitted findings in multi-pathology volumes.}
\label{fig:gpt4o_errors}
\end{figure}

\subsection{Prompt Templates}
\label{sec:prompt_templates}

\begin{figure}[h!]
\centering
\includegraphics[width=\textwidth]{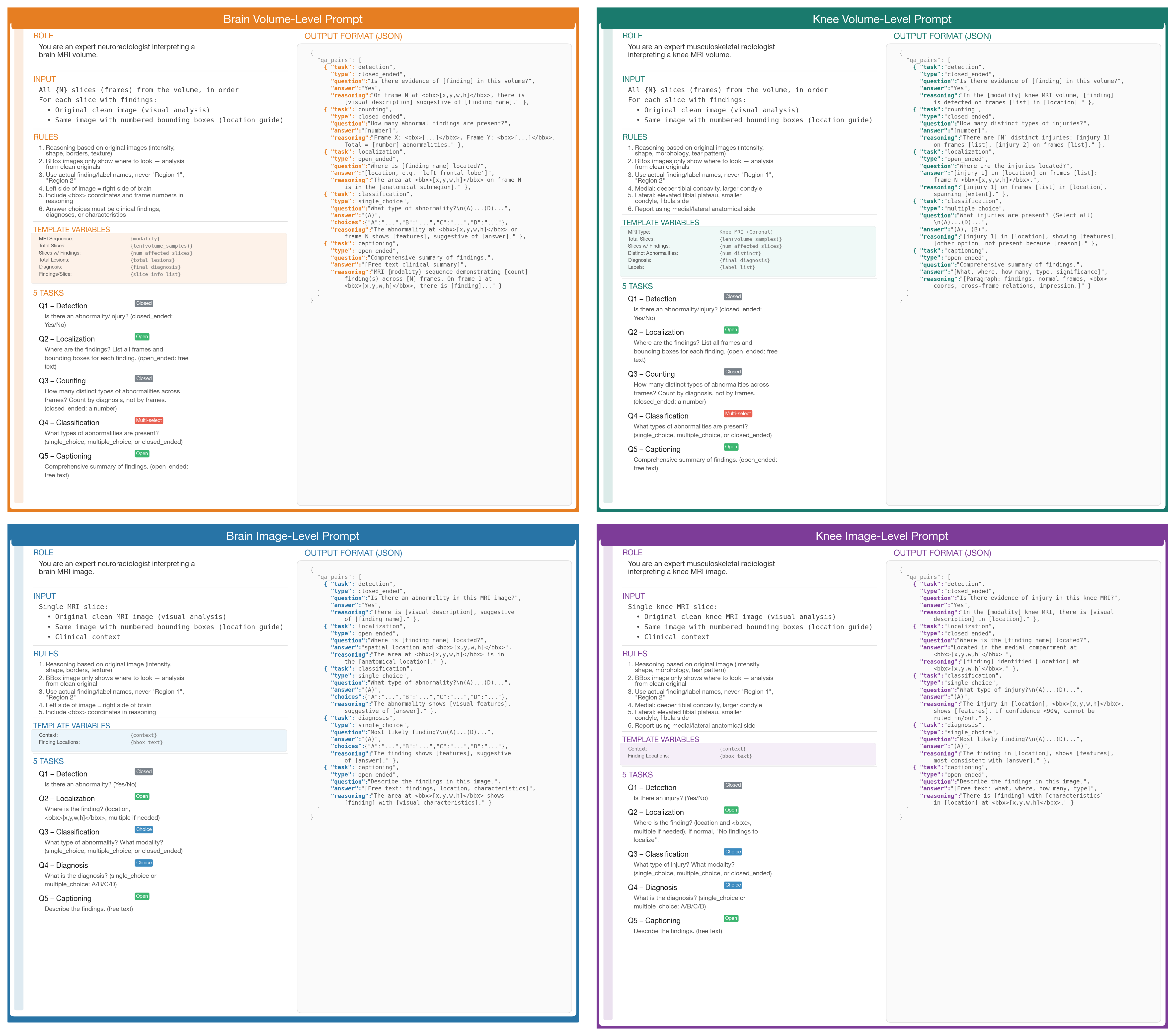}
\caption{\textbf{Overview of the four GPT-4o prompt templates used for QA pair generation.} Each card summarizes the role, input format, rules, template variables, five hierarchical tasks, reasoning and expected JSON output schema for one of the four prompt types: brain volume-level, knee volume-level, brain image-level, and knee image-level.}
\label{fig:prompt_cards}
\end{figure}

\end{document}